\pdfoutput=1

\documentclass[11pt]{article}
\usepackage{amsmath}
\usepackage{amssymb}
\usepackage[final]{acl}

\usepackage{times}
\usepackage{latexsym}

\usepackage[T1]{fontenc}

\usepackage[utf8]{inputenc}

\usepackage{microtype}

\usepackage{inconsolata}

\usepackage{graphicx}

\usepackage{enumitem}  
\usepackage{subcaption}

\usepackage{amsmath}
\usepackage{booktabs}
\usepackage{hyperref}

\usepackage[marginal]{footmisc}

\title{Related Knowledge Perturbation Matters: Rethinking Multiple Pieces of Knowledge Editing in Same-Subject}

\author{
 \textbf{Zenghao Duan}\textsuperscript{1,2,*},
 \textbf{Wenbin Duan}\textsuperscript{3,*},
 \textbf{Zhiyi Yin}\textsuperscript{1,\(\dagger\)},
 \textbf{Yinghan Shen}\textsuperscript{1,\(\dagger\)},
\\
 \textbf{Shaoling Jing}\textsuperscript{1},
 \textbf{Jie Zhang}\textsuperscript{1},
 \textbf{Huawei Shen}\textsuperscript{1},
 \textbf{Xueqi Cheng}\textsuperscript{1},
\\
 \textsuperscript{1}Institute of Computing Technology, Chinese Academy of Sciences, Beijing, China
 \\
 \textsuperscript{2}University of Chinese Academy of Sciences, Beijing, China
 \\
 \textsuperscript{3}People's Public Security University of China, Beijing, China
\\
 \small{
     \href{mailto:email@domain}{\{duanzenghao24s, yinzhiyi, Shenyinghan, jingshaoling, zhangjie, shenhuawei, cxq\}@ict.ac.cn}
 }
}

\begin{document}

\maketitle

\setcounter{footnote}{0} 

\stepcounter{footnote}
\footnotetext[\value{footnote}]{* Equal Contributions}
\stepcounter{footnote}
\footnotetext[\value{footnote}]{\(\dagger\) Corresponding authors}

\setcounter{footnote}{1} 
\footnotetext[1]{Our benchmark and source code are available at: \url{https://github.com/Zhow01/S2RKE}}

\begin{abstract}



Knowledge editing has become a promising approach for efficiently and precisely updating knowledge embedded in large language models (LLMs). In this work, we focus on \textbf{Same-Subject Editing}, which involves modifying multiple attributes of a single entity to ensure comprehensive and consistent updates to entity-centric knowledge. Through preliminary observation, we identify a significant challenge: \textit{Current state-of-the-art editing methods struggle when tasked with editing multiple related knowledge pieces for the same subject.} To address the lack of relevant editing data for identical subjects in traditional benchmarks, we introduce the \textbf{$\text{S}^2\text{RKE}$} (\textbf{S}ame-subject \textbf{R}elated \textbf{K}nowledge \textbf{E}diting) benchmark. Our extensive experiments reveal that only mainstream locate-then-edit methods, such as ROME and MEMIT, exhibit \textit{"related knowledge perturbation,"} where subsequent edits interfere with earlier ones. Further analysis reveals that these methods over-rely on subject information, neglecting other critical factors, resulting in reduced editing effectiveness.

\end{abstract}

\section{Introduction}



The dynamic nature of real-world knowledge necessitates efficient methods for updating specific facts in large language models (LLMs) \cite{achiam2023gpt, touvron2023llama} without compromising their overall performance.
\textit{Knowledge editing}(a.k.a., \textit{model editing}) \cite{yao-etal-2023-editing} has emerged as a promising solution to address this challenge, enabling targeted updates to model parameters without requiring full retraining. 
Among existing methods, \textit{locate-then-edit} methods, such as ROME \cite{meng2022locating} and MEMIT \cite{meng2022memit}, have shown effectiveness in making precise modifications to Transformer layer parameters \cite{vaswani2017attention}. However, their broader applicability across diverse editing scenarios remains insufficiently explored.

\begin{figure}[t]
  \center
  \includegraphics[width=\columnwidth]{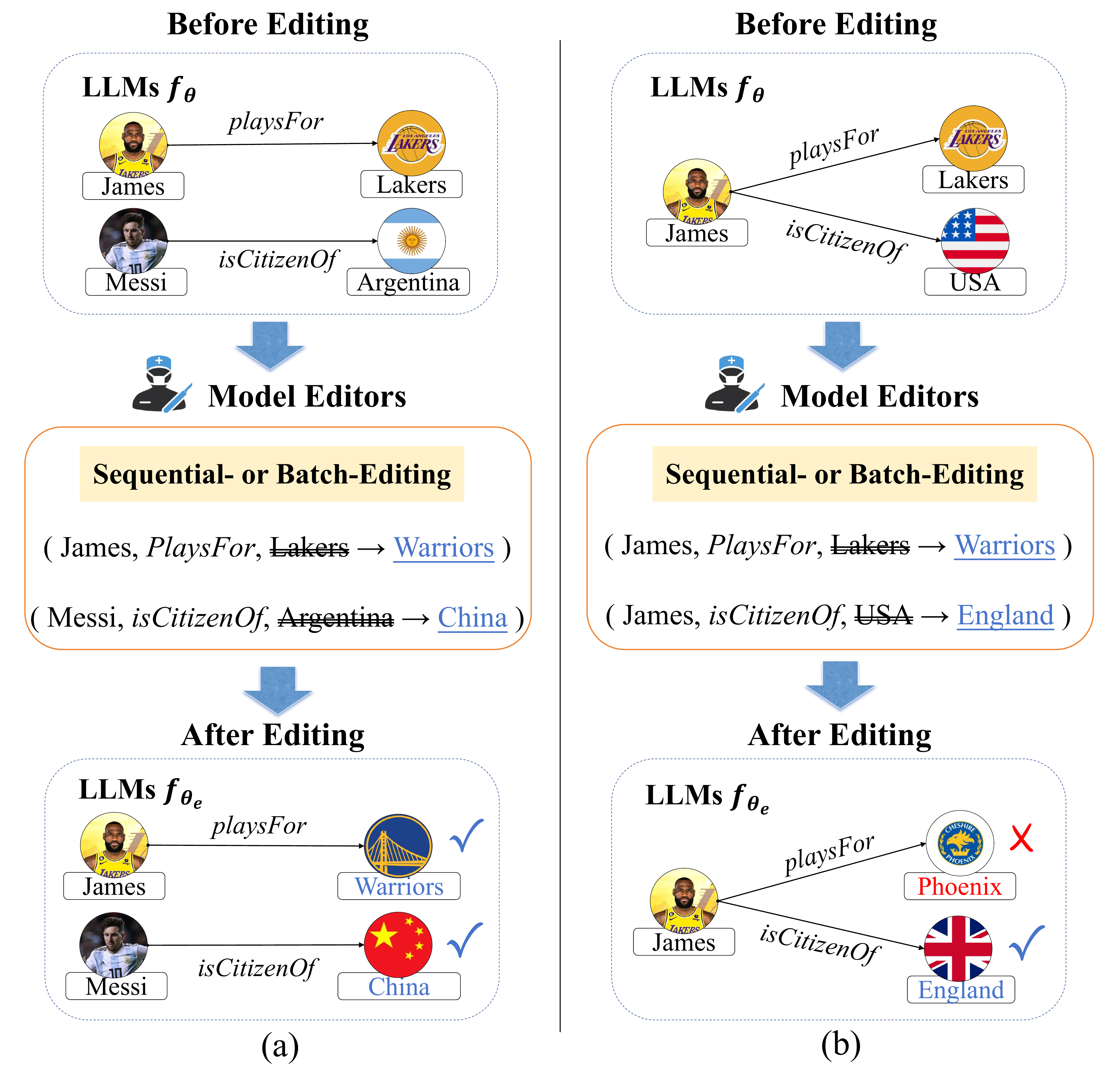}
  \caption{Comparison of performance on Different and Same-Subject Editing. (a) Editing individual knowledge pieces for distinct subjects, "James" and "Messi," results in excellent performance. (b) Editing two related knowledge pieces for the same subject, "James," leads to poor performance.}
  \label{fig:introduction}
\end{figure}

\begin{figure*}[t]
  \centering
  \includegraphics[width=2\columnwidth]{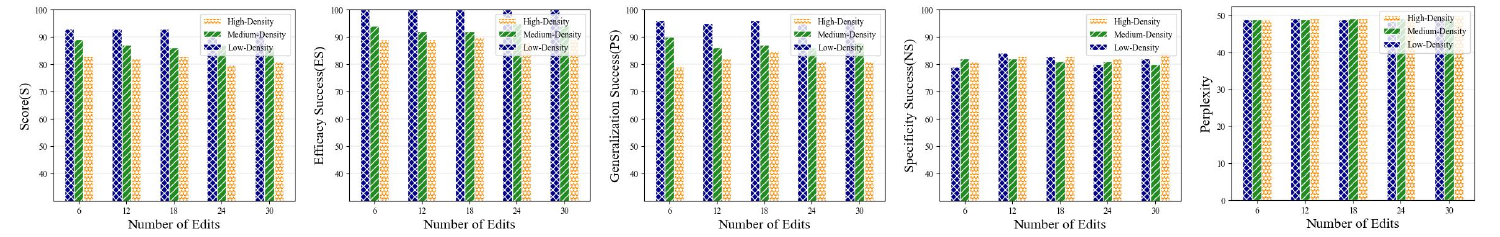}
  \caption{The results of \textit{sequential-editing} by three different schemes on GPT-J using MEMIT, comparing five evaluation metrics. The values of Score(\textbf{S}), Efficacy Success(\textbf{ES}) and Paraphrase Success(\textbf{PS}) always decreased with the subject density, but Neighborhood Success(\textbf{NS}) and Perplexity(\textbf{PPL}) remained unchanged.}
  \label{fig:pilot sequential}
\end{figure*}


In particular, \textbf{Same-Subject Editing}, modifying multiple attributes of a single entity, plays a critical role in ensuring comprehensive and consistent updates to entity-centric knowledge. 
As shown in Figure~\ref{fig:introduction}, an entity like "James" may require simultaneous edits to attributes such as "isCitizenOf," "playsFor," and others. 
This process refines the entity's representation by resolving attribute conflicts and synchronizing interdependent facts. Despite its significance, same-subject editing has largely been overlooked in existing research.

Through preliminary observations, we identify an unusual failure: \textit{Some top-performing editing methods struggle to edit multiple related knowledge pieces for the same subject.} As illustrated in Figure~\ref{fig:introduction}, model editors perform well when editing individual knowledge pieces for different subjects, such as "James" and "Messi" (Figure~\ref{fig:introduction}a). However, when tasked with editing two related pieces of knowledge for the same subject, "James," these editors become significantly less effective (Figure~\ref{fig:introduction}b). This observation raises two key questions:

\begin{itemize}[itemsep=0pt, parsep=0pt]
    \item \textit{Is this failure a common issue across different LLMs and editing methods?}
    \item \textit{What causes the failure when editing multiple related knowledge pieces about same subject?}
\end{itemize}

Existing benchmarks, such as COUNTERFACT \cite{meng2022locating}, lack sufficient examples of same-subject editing, making it difficult to explore the underlying mechanisms of this failure. To address this gap, we introduce the \textbf{$\text{S}^2\text{RKE}$} (\textbf{S}ame-subject \textbf{R}elated \textbf{K}nowledge \textbf{E}diting) benchmark, which associates each subject with multiple related edits. We systematically evaluate various editing methods on LLMs of different sizes using $\text{S}^2\text{RKE}$, applying both \textit{sequential-editing} and \textit{batch-editing}. Surprisingly, the results show that only mainstream locate-then-edit methods, such as MEMIT \cite{meng2022memit}, fail to effectively update multiple related information for  the same subject. Moreover, our in-depth analysis reveals that this failure occurs because subsequent edits interfere with previous ones, a phenomenon we term \textit{"related knowledge perturbation."}

Furthermore, we find that locate-then-edit methods exhibiting \textit{"related knowledge perturbation"} update the weight matrix of the MLP module by calculating key-value pairs. Specifically, the key is derived from the input of the subject’s last token in the MLP module’s down-sampling layer. 
Our experiments conclude that the perturbation arises from an over-reliance on subject information during editing. When multiple related pieces of knowledge share the same subject, the calculated keys remain highly similar. As a result, subsequent edits interfere with earlier ones, diminishing the overall effectiveness of the editing process.

In essence, our main contributions are as follows: (1) We propose the $\text{S}^2\text{RKE}$ benchmark for Same-Subject Editing and highlight the issue of "\textit{related knowledge perturbation}." (2) We demonstrate that locate-then-edit methods fail to update multiple related facts for the same subject due to an over-reliance on subject-specific information.

\begin{figure*}[t]
  \centering
  \includegraphics[width=2\columnwidth]{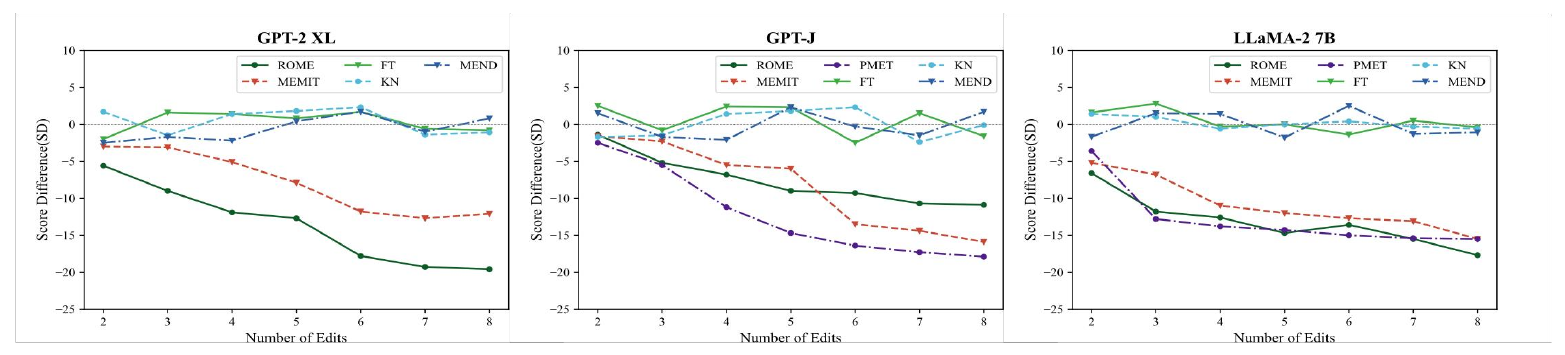}
  \caption{The results of differences in  \textit{sequential-editing} results in two scenarios on three LLMs by six editing methods. \textbf{Score Difference (SD)} represents the difference in editing performance between the two experimental schemes when editing the same amount of knowledge under the same method.}
  \label{fig:Failue sequential}
\end{figure*}

\section{Preliminary}
\subsection{Knowledge in Language Model}

Autoregressive, decoder-only large language models (LLMs) process a sequence of tokens \(x = [x_1, \dots, x_T] \in X\), where each token \(x_i \in V\) is drawn from a vocabulary \(V\), and predict the probability distribution \(y \in Y \subset \mathbb{R}^{|V|}\) for the next token in the sequence. In the Transformer architecture, each token \(x_i\) is embedded into a sequence of hidden state vectors \(h^{(l)}_i\), starting from the initial embedding \(h^{(0)}_i = \text{emb}(x_i) + \text{pos}(i) \in \mathbb{R}^H\). The final output \(y = \text{decode}(h^{(L)}_T)\) is derived from the last hidden state of the sequence. At each layer \(l\), the hidden state \(h_i^{(l)}\) is updated through a combination of global attention \(a_i^{(l)}\) and contributions from the local MLP \(m_i^{(l)}\), where each token attends only to preceding tokens:

\begin{equation}
h_i^{(l)} = h_i^{(l-1)} + a_i^{(l)} + m_i^{(l)},
\end{equation}
\begin{equation}
a_i^{(l)} = \text{attn}^{(l)} \left( h_1^{(l-1)}
h_2^{(l-1)}, \dots, h_i^{(l-1)} \right),
\end{equation}
\begin{equation}
m_i^{(l)} = W_{\text{proj}}^{(l)} \sigma \left( W_{\text{fc}}^{(l)} \gamma \left( a_i^{(l)} + h_i^{(l-1)} \right) \right),
\end{equation}

In many previous studies, knowledge has been represented as triples $(s, r, o)$, where $s$, $r$, and $o$ denote subject, relation, and object respectively (e.g., \text{James} ($s$), \text{playsFor} ($r$), and \text{Lakers} ($o$))\cite{meng2022locating,li2024pmet}. Researchers typically design natural language templates for each relation type and integrate these templates with subject terms to create question or cloze-style prompts.

\subsection{Same-Subject Editing}


In a broad sense, knowledge editing aims to allow for the querying and modification of a wide range of facts within language models by combining different subjects \((s)\) and relations \((r)\) as prompts. Existing work typically focuses on modifying individual facts expressed as \((s, r, o) \rightarrow (s, r, o_{\ast})\), where each subject \((s)\) is associated with a specific relation \((r)\). However, traditional editing often isolates the editing process to a single relation. This leads to the discontinuation of further knowledge edits for the same subject and a shift towards editing knowledge for a new subject. It risks overlooking potential perturbations in knowledge when editing multiple related facts for the same subject.


We introduce the concept of  \textbf{Same-Subject Editing}, where multiple relations are edited simultaneously for a single subject. Instead of focusing solely on the traditional \((s, r, o)\) format, we extend the editing process to structured prompts such as \((s, R, O)\), where $R = \{r_i\}_{i=1}^N$ represents a set of relations and $O = \{o_i\}_{i=1}^N$ represents their corresponding objects. For example, \{(\text{"James"}, \text{"playsFor"}, \text{"Lakers"}), (\text{"James"}, \text{"isCitizenOf"}, \text{"USA"})\}.
We formally define the edited fact set as $e = {(s, r_i, o_i)}_{i=1}^N$ and define the edited model as $M^* = F(M, e)$, where $F$ is the editing function that updates the original model $M$. It ensures that knowledge updates remain consistent across all related attributes of the same subject.


\section{Pilot Observation}
In this section, we conduct a pilot observation to reveal potential issues with same-subject editing.



\textbf{Evaluation Setup.} We focus on using MEMIT \cite{meng2022memit} to edit GPT-J \cite{wang2021gpt}, since their excellent performance in editing multiple pieces of knowledge. To analyze the impact of editing density—defined here as the average number of related edits per subject in the editing sequence—we divide our experimental schemes into three categories:

\begin{enumerate}[label=\alph*), itemsep=0pt, parsep=0pt]
    \item \textit{\textbf{High-Density}: Edit \textbf{n} pieces of knowledge in total, with each subject edited for \textbf{3} related pieces of knowledge.}
    \item \textit{\textbf{Medium-Density}: Edit \textbf{n} pieces of knowledge in total, with each subject edited for \textbf{2} related pieces of knowledge.}
    \item \textit{\textbf{Low-Density}:  Edit \textbf{n} pieces of knowledge in total, with each subject edited for \textbf{1} related pieces of knowledge.}
\end{enumerate}


Based on the above schemes, we select qualified data from COUNTERFACT \cite{meng2022locating} and conduct experiments using both \textit{sequential-editing} and \textit{batch-editing} (See Appendix~\ref{sec:comparison} for comparison of sequential- and batch-editing). The editing performance is comprehensively evaluated across four dimensions: \textbf{efficacy}, \textbf{generalization}, \textbf{specificity}, and \textbf{overall performance} (See Appendix \ref{Evolution metrics} for detailed metric descriptions).

\textbf{Result \& Analysis.} Figure~\ref{fig:pilot sequential} and Figure~\ref{fig: pilot batch} show the experimental results of employing MEMIT to edit GPT-J through \textit{sequential-editing} and \textit{batch-editing}, respectively.   
It is evident that when editing the same number of knowledge, the denser the subject distribution, the worse the editing performance, while the impact on the model's downstream performance remains similar.
However, the scarcity of sufficiently dense same-subject instances in existing editing datasets limits the scope of experimental verification. We will further investigate this phenomenon in subsequent sections.




\section{Related Knowledge Perturbation}


Furthermore, we construct a benchmark and evaluate the performance of editing methods when editing related knowledge for the same subject.

\subsection{$\text{S}^2\text{RKE}$ Benchmark}


 We introduce the \textbf{$\text{S}^2\text{RKE}$} (\textbf{S}ame-subject \textbf{R}elated \textbf{K}nowledge \textbf{E}diting) benchmark, specifically designed to facilitate the editing of multiple related pieces of knowledge for each subject. It covers six categories of subjects, comprising of 4,503 subjects and 43 relationships, with each subject having an average of 4.9 related knowledge items. See Appendix \ref{technical details} for additional technical details about its construction and Table \ref{table: Comparison of different benchmarks} for comparison of statistics between $\text{S}^2\text{RKE}$ and COUNTERFACT.

\begin{table}[t]
\centering
\scalebox{0.73}{
\begin{tabular}{lcc}
\hline
\textbf{Item}  & \textbf{$\text{S}^2\text{RKE}$}& \textbf{COUNTERFACT} \\
\hline
Records       & 22064   & 21919   \\
Subjects      & 4503    & 20391  \\
Relations     & 43      & 32   \\
Maximum records per subject    & 13     & 4   \\
Minimum records per subject    & 3      & 1   \\
Average records per subject    & 4.9    & 1.1  \\
\hline
\end{tabular}}
\caption{Comparison of different benchmarks.}
\label{table: Comparison of different benchmarks}
\end{table}



\subsection{Failure of Editing Methods}


\textbf{Editing Methods.} We evaluate six widely-used editing methods: ROME \cite{meng2022locating}, MEMIT \cite{meng2022memit}, PMET \cite{li2024pmet}, FT \cite{zhu2021modifying}, MEND \cite{mitchell2022fast}, and KN \cite{dai-etal-2022-knowledge}. 

\noindent\textbf{Selected LLMs.} Experiments are conducted on three LLMs with different parameter sizes: GPT-2 XL (1.5B) \cite{radford2019language}, GPT-J (6B) \cite{wang2021gpt}, and LLaMA-2 (7B) \cite{touvron2023llama}.



We design two experimental schemes to assess how editing related knowledge impacts performance: \textbf{\textit{Same-Subject}}, where all edited knowledge shares the same subject, \textbf{\textit{Different-Subject}}, where each edit involves a different subject. Experimental data are selected from the $\text{S}^2\text{RKE}$ benchmark. 

Our pilot observation indicates that while knowledge correlation impacts editing effectiveness, it has little effect on overall model performance. So we focus on the \textbf{Score(S)} metric and introduce the \textbf{Score Difference (SD)} metric, defined as SD = Score(same-subject) – Score(different-subject), to quantify performance degradation when editing related knowledge for the same subject. To ensure reliability, each test was repeated 30 times with different editing instances. See Appendix \ref{detail experiment} for more details.




\textbf{Result \& Analysis.} Figure~\ref{fig:Failue sequential} and Figure~\ref{fig:Failue batch} show the results of \textit{sequential-editing} and \textit{batch-editing} on three LLMs using six methods, respectively. The line in each figure represents the Score Difference (SD). The results show that locate-then-edit methods (e.g., ROME, MEMIT, PMET) suffer significant performance degradation under Same-Subject editing, as reflected by a substantial negative Score Difference (SD). In contrast, methods with generally lower editing effectiveness show minimal sensitivity to the relatedness of the edited knowledge. These findings confirm that knowledge correlation markedly impairs the editing performance of certain methods.



\subsection{Analysis of Failures}

We further examine how the sequence of knowledge edits affects locate-then-edit methods by isolating the interference of sequential updates. For this purpose, we devised two experimental settings: \textbf{\textit{Homogeneous-Editing}}, where the first and last edits target the same subject, and \textbf{\textit{Heterogeneous-Editing}}, in which they target different subject. Experiments were performed using ROME, MEMIT, and PMET across three LLMs, with each configuration repeated 30 times on different instances from the $\text{S}^2\text{RKE}$ benchmark to ensure robust results. 


\begin{figure}[t]
  \centering
  \includegraphics[width=\columnwidth]{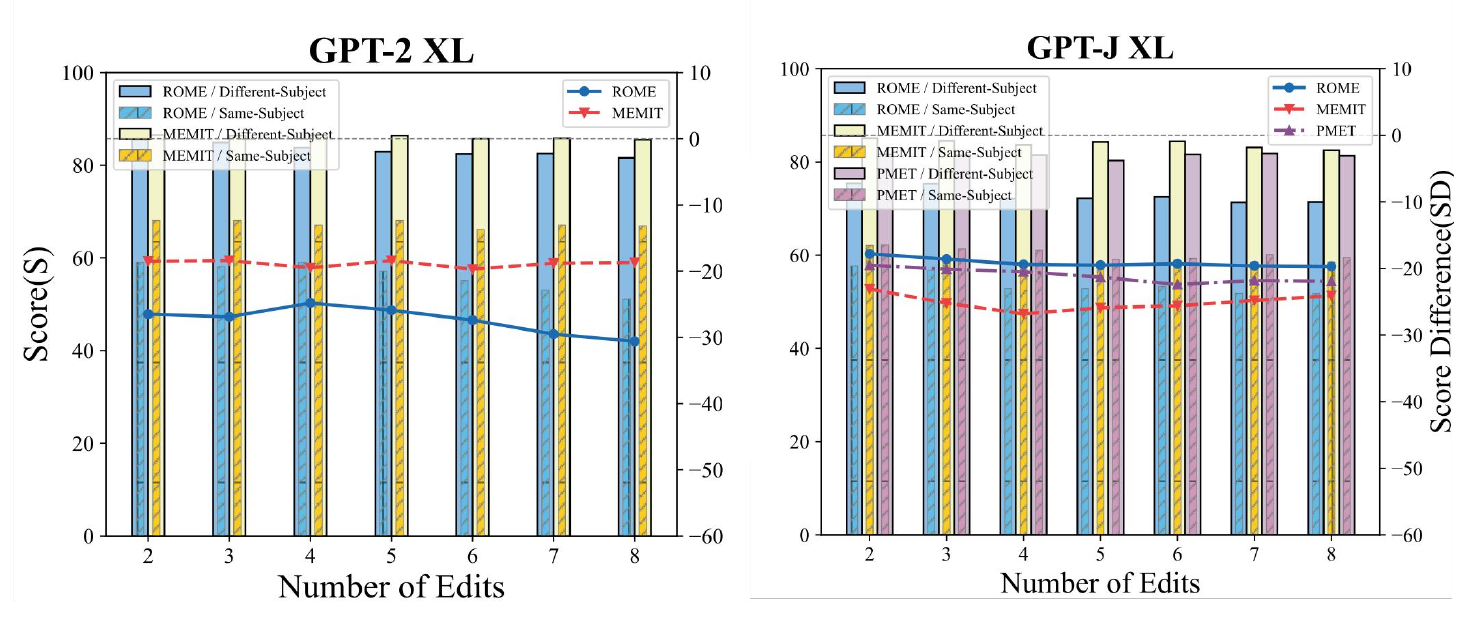}
  \caption{The results of \textit{sequential-editing} on GPT-2 XL and GPT-J using mainstream locate-then-edit methods. The bars represent the \textbf{Score (S)} of two strategies, and the line represents the \textbf{Score Difference (SD)} between the two strategies.}
  \label{fig:defference edit 1}
\end{figure}

\textbf{Result \& Analysis.} Figure~\ref{fig:defference edit 1} shows the sequential-editing results on GPT-2 XL and GPT-J, while Figures~\ref{fig:defference edit 2} and~\ref{fig: Failue of defference batch} provide additional results. Under the Homogeneous-Editing setting, the initial edit's score is much lower than in the Heterogeneous-Editing condition. This clearly indicates that later edits interfere with earlier ones. We call this effect "\textit{related knowledge perturbation}," which exposes a key limitation of current locate-then-edit approaches when processing multiple sequential updates. These findings highlight the need for better strategies in managing sequential knowledge updates. The next section will analysis the causes of \textit{related knowledge perturbation}.


\section{Perturbation Analysis}



\begin{figure}[t]
  \centering
  \includegraphics[width=\columnwidth]{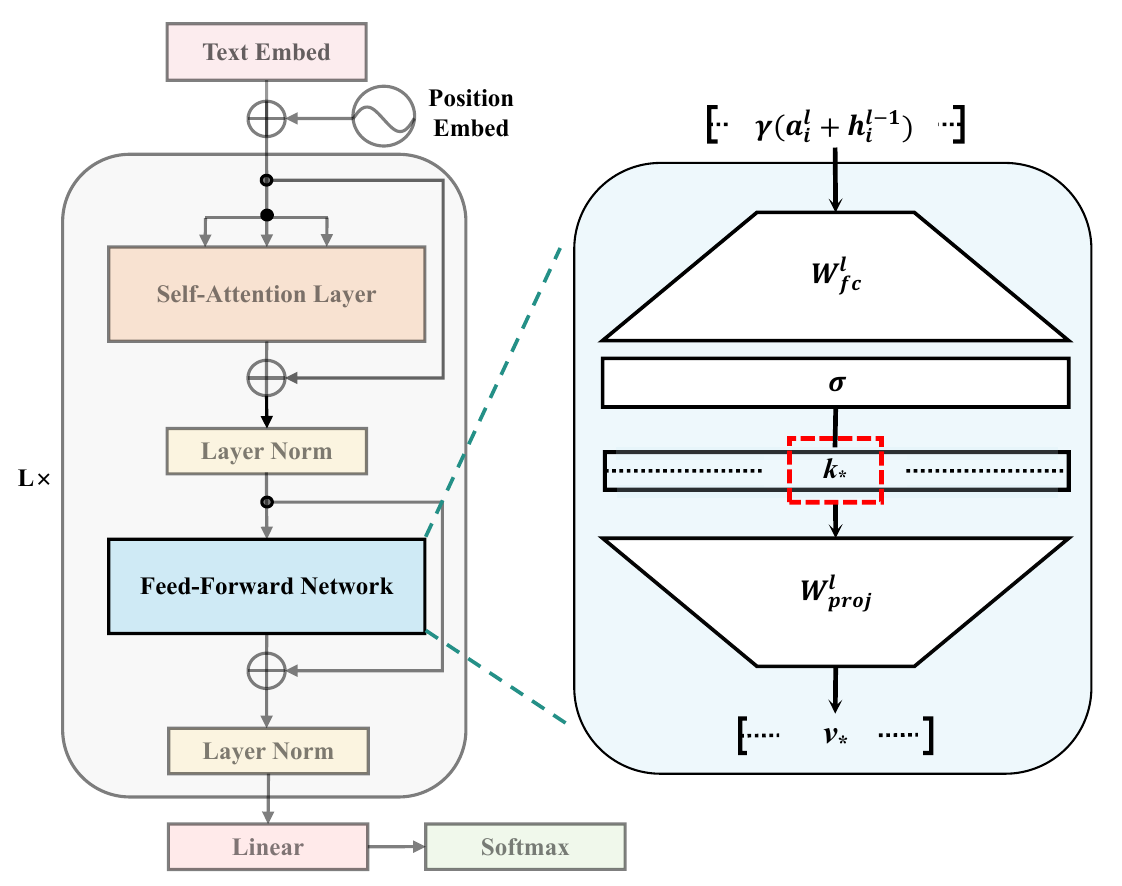}
  \caption{Illustration of related knowledge perturbation in same-subject editing.}
  \label{fig:illustration}
\end{figure}

\subsection{Causes of Perturbation}

Our experiments show that only mainstream locate-then-edit methods (e.g., ROME and MEMIT) exhibit \textit{related knowledge perturbation}. These methods all employ causal tracing to identify that factual knowledge is primarily stored in the early MLP layers of LLMs. Based on the hypothesis that "the MLP modules in Transformer layers can be viewed as linear key-value associative memory," \cite{geva2020transformer} they solve for $Wk=v$, where $W$ represents the downsampling component $W_{\text{proj}}^{(l)}$ of MLP, and the key-value pair $(k, v)$ corresponds to a factual triplet $t=(s, r, o)$, as shown in Figure~\ref{fig:illustration}. Here, $k$ represents the subject $s$, while $v$ encodes the attributes of $s$, including $r$ and $o$. To update \(t\) to \(t_{\ast}=(s, r, o_{\ast})\), they compute a new key \(k_{\ast}\) and value \(v_{\ast}\) via an update \(\Delta W\).

However, $k_{\ast}$ is only derived from the input of the subject's last token in the MLP module's downsampling layer:

\begin{equation}
  \label{eq:example}
  k_{\ast} = \frac{1}{N} \sum_{i=1}^{N} \mathcal{K}(x_i \oplus p),
\end{equation}

\noindent where $\mathcal{K}$ is the output of the first MLP layer in transformer block, $x_i$ represents the randomly sampled prefixes, and $\oplus$ denotes the string concatenation operator. 


Therefore, we speculate that \textit{"related knowledge perturbation"} stems from an over-reliance on subject information. When editing multiple pieces of knowledge for the same subject $s$, the key value $k_{\ast}$ remains constant, causing later edits to interfere with earlier ones and reducing performance.

\subsection{Experiment Validation}

To verify the above speculation, we used MEMIT to edit two pieces of knowledge on GPT-J through \textit{sequential-editing} and \textit{batch-editing}, designing two experimental schemes: \textbf{\textit{Same-Subject}} and \textbf{\textit{Different-Subject}}. We then examine the relationship between the \textbf{cosine similarity} of the two keys and the \textit{Efficacy Success} of editing the first piece of knowledge. Cosine similarity was chosen because it measures how similar the two keys are in vector space, helping us understand how closely related the two knowledge pieces are.


\begin{figure}[t]
  \centering
  \includegraphics[width=\columnwidth]{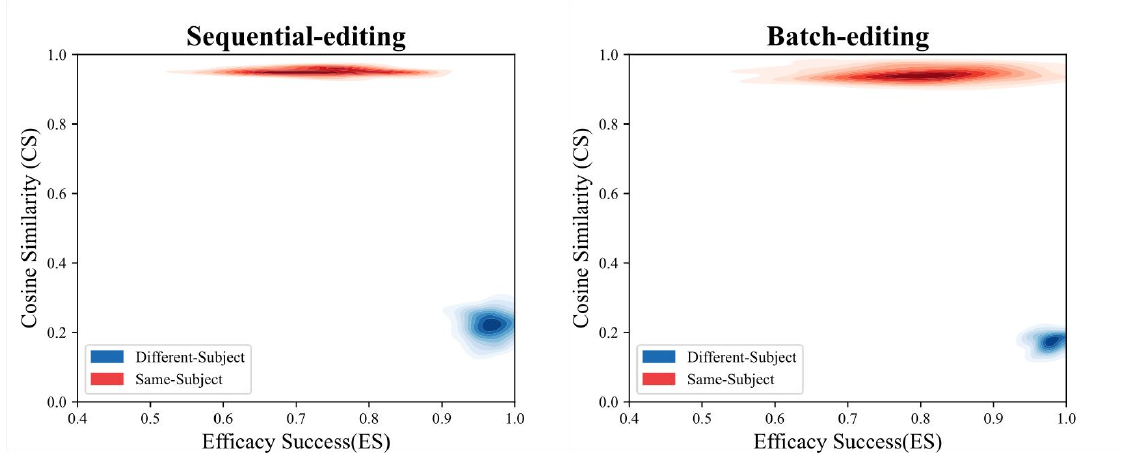}
  \caption{The relationship between the \textbf{cosine similarity} of keys and the \textbf{Efficacy Success (ES)} of the first knowledge editing using MEMIT to edit GPT-J, under \textit{sequential-editing} and \textit{batch-editing}. }
  \label{fig:similar}
\end{figure}

\textbf{Result \& Analysis}\quad Figure \ref{fig:similar} shows the relationship between key similarity and the first knowledge editing Efficacy Success. The results indicate that when two pieces of knowledge related to the same subject are edited, the CS of the key approaches 1. Meanwhile, the ES of editing the first piece of knowledge is significantly lower compared to the case where the two edited pieces of edited knowledge are related to different subjects. This supports our hypothesis that since the key calculation only focuses on subject information, subsequent edits for the same subject interfere with earlier ones, leading to \textit{"related knowledge perturbation"}.




\section{Conclusion}
In this paper, we identify a key limitation of mainstream locate-then-edit methods, called \textit{"related knowledge perturbation"}, which occurs when editing multiple related pieces of knowledge for the same subject. Using the $\text{S}^2\text{RKE}$ benchmark, we show through experiments that over-reliance on subject information leads to interference between subsequent edits, highlighting the challenges in same-subject editing. 

\section{Limitation}

We acknowledge several limitations in our work. First, while this paper provides an initial exploration into the complex correlations between knowledge and identifies the phenomenon of related knowledge perturbation, it does not propose a comprehensive solution to address this issue. This omission leaves room for future research to develop effective mitigation strategies. 

Additionally, due to computational resource constraints, our experiments did not extend to larger language models, such as Llama2-13b. Future investigations could benefit from testing our findings on such models to further validate the effectiveness and generalizability of the observed phenomena.

\section{Acknowledgement}
We would like to express our sincere gratitude to all the reviewers for their valuable feedback, which greatly contributed to the improvement of this research. This work was supported by the Strategic Priority Research Program of the Chinese Academy of Sciences (No. XDB0680202).

\bibliography{custom}
\appendix

\section{Related Works}  \label{sec:appendix}

\subsection{Knowledge Editing}
Model editing has gained significant attention for its ability to efficiently update LLMs. Existing approaches can be categorized into four types: \textbf{Fine-tuning} mainly applies layer-wise adjustments to incorporate new knowledge into LLMs \cite{zhu2021modifying}. \textbf{Meta Learning} trains hypernetworks to act as editors, predicting parameter updates to inject new knowledge \cite{de2021editing, mitchell2022fast}. \textbf{Memory-based} enhances LLMs with external memory or additional parameters, allowing new knowledge to be added without altering LLMs \cite{mitchell2022memory, huang2023transformerpatcher}.

Among all types, \textbf{Locate-then-Edit} has gained significant traction for its ability to modify specific knowledge within LLMs. Methods like KN\cite{dai-etal-2022-knowledge} and ROME\cite{meng2022locating} locate and update factual knowledge by targeting neurons or multi-layer perceptrons (MLPs) that store such information. MEMIT\cite{meng2022memit} extends ROME by distributing updates across multiple intermediate MLP sublayers, enabling large-scale knowledge editing. Additionally, PMET\cite{li2024pmet} combines information from both multi-head Self-attention (MHSA) and MLP modules during optimization, producing more accurate MLP outputs for final edits.

While model editing has shown great promise, some researches have identified issues such as model collapse\cite{yang2024butterfly, gu2024model} and knowledge conflicts\cite{li2024unveiling}. This paper focuses on how the correlation between knowledge impacts the performance of model editing, particularly in the context of multiple knowledge edits.

\subsection{Sequantial-editing vs. Batch-editing} \label{sec:comparison}

\textit{Sequential-editing} and \textit{batch-editing} are two strategies commonly used to update large amounts of knowledge in LLMs\cite{yao-etal-2023-editing}. Specifically, \textit{sequential-editing} refers to making multiple edits one after another, where the model should ideally retain previous changes as new edits are introduced. In contrast, \textit{batch-editing} involves editing multiple pieces of knowledge in a model at once. Notably, these two strategies can be combined to create a more flexible knowledge editing approach.

For the purposes of this study, we evaluate these strategies independently: In \textit{sequential-editing}, the batch size is set to 1, and in \textit{batch-editing}, the number of consecutive edits is set to 1, ensuring clear comparisons and facilitate experimental evaluation.


\section{Details of $\text{S}^2\text{RKE}$ Benchmark}  \label{technical details}

\subsection{Data Construction}

In this paper, \textbf{$\text{S}^2\text{RKE}$} (\textbf{S}ame-subject \textbf{R}elated \textbf{K}nowledge \textbf{E}diting) benchmark is built on the YAGO3.0.3, which combines Wikipedia, WordNet, GeoNames and other data sources, and was released in 2022. The construction process is detailed below, covering four key aspects:

\textbf{Triple filtering.}\quad Based on YAGO's top-level classification, we categorize the entities to be edited into six groups: Person, Building, Organization, Abstraction, Artifact and GeoEntity. From these categories, we screen out 43 relationships. Unlike COUNTERFACT, $\text{S}^2\text{RKE}$ innovatively includes both literal- and data-type relationships, enabling broader coverage of relationship types. Finally, We then select entities with the most relationship instances from each category and generated correct triplets $(s, r, o)$.

\textbf{Requested rewrite.}\quad To evaluate model efficacy, we select the relation $r$ from the triplet $(s, r, o)$ and generate a counterfactual triplet $(s, r, o_{\ast})$. We create natural language templates $P(r)$ for each relation $r$, using ChatGPT-4o to generate templates based on examples from the PARAREL \cite{elazar2021measuring} dataset. After generating multiple templates, we manually select the three most suitable ones to ensure test diversity and template consistency.

\textbf{Paraphrase prompts.}\quad To evaluate the generalization of model editing methods, we use the moonshot-v1 for generating longer text, combined with the description of the edited entity and a simplified prompt template for each relation. This process produce semantically equivalent but more complex sentences $P^P$, designed to test the model’s ability to handle diverse expressions.

\textbf{Neighborhood prompts.}\quad In order to evaluate the specificity of the model editing methods, we identify related triples $(s_{\ast}, r_{\ast}, o)$ for the object $o$ of the original triplet $(s, r, o)$, using the YAGO database. These neighborhood triplets are converted into natural language $P^N$ using simple templates$P(r_{\ast})$, specifically constructed for each relation $r_{\ast}$.

\begin{table}[t]
\centering
\scalebox{0.75}{
\begin{tabular}{lcccc}
\hline
\textbf{Categories}  & \textbf{Subjects} & \textbf{Relations} & \textbf{Edits(all)} & \textbf{Edits(Avg)} \\
\hline
Person      & 592   & 29 & 5706  & 9.6 \\
Organization& 874   & 7  & 2897   & 3.3 \\
Building    & 679   & 6  & 3419  & 4.6 \\
Artifact    & 857   & 6  & 3632  & 4.2 \\
Abstraction & 734   & 8  & 2203  & 3.0 \\
GeoEntity   & 912   & 12 & 4207  & 5.0 \\
\hline
All         & 4503  & 43 & 22064 & 4.9 \\
\hline
\end{tabular}}
\caption{Data statistics of the $\text{S}^2\text{RKE}$ benchmark.}
\label{table:MAPKE}
\end{table}

\begin{table*}[t]
\centering
\small
\scalebox{0.8}{ 
\begin{tabular}{llll}
\hline
\textbf{ID} & \textbf{Relation} & \textbf{Domain} & \textbf{Range} \\
\hline
1  & \texttt{<hasPages>}           & rdfs:domain owl:Thing                               & rdfs:range xsd:nonNegativeInteger \\
2  & \texttt{<isCitizenOf>}        & rdfs:domain \texttt{<wordnet\_person\_100007846>}   & rdfs:range \texttt{<wordnet\_country\_108544813>} \\
3  & \texttt{<diedOnDate>}         & rdfs:domain \texttt{<wordnet\_person\_100007846>}   & rdfs:range xsd:date \\
4  & \texttt{<hasGender>}          & rdfs:domain \texttt{<wordnet\_person\_100007846>}   & rdfs:range \texttt{<wordnet\_sex\_105006698>} \\
5  & \texttt{<wasBornOnDate>}      & rdfs:domain \texttt{<wordnet\_person\_100007846>}   & rdfs:range xsd:date \\
6  & \texttt{<hasDuration>}        & rdfs:domain owl:Thing                               & rdfs:range \texttt{<s>} \\
7  & \texttt{<hasWeight>}          & rdfs:domain \texttt{<wordnet\_physical\_entity\_100001930>} & rdfs:range \texttt{<kg>} \\
8  & \texttt{<hasHeight>}          & rdfs:domain \texttt{<wordnet\_physical\_entity\_100001930>} & rdfs:range \texttt{<m>} \\
9  & \texttt{<hasLength>}          & rdfs:domain \texttt{<yagoGeoEntity>}                & rdfs:range \texttt{<km>} \\
10 & \texttt{<hasWonPrize>}        & rdfs:domain \texttt{<yagoLegalActorGeo>}            & rdfs:range \texttt{<wordnet\_award\_106696483>} \\
11 & \texttt{<owns>}               & rdfs:domain \texttt{<yagoLegalActorGeo>}            & rdfs:range owl:Thing \\
12 & \texttt{<created>}            & rdfs:domain \texttt{<yagoLegalActor>}               & rdfs:range owl:Thing \\
13 & \texttt{<participatedIn>}     & rdfs:domain \texttt{<yagoLegalActorGeo>}            & rdfs:range owl:Thing \\
14 & \texttt{<isAffiliatedTo>}     & rdfs:domain \texttt{<yagoLegalActor>}               & rdfs:range \texttt{<wordnet\_organization\_108008335>} \\
15 & \texttt{<hasAcademicAdvisor>}  & rdfs:domain \texttt{<wordnet\_person\_100007846>}   & rdfs:range \texttt{<wordnet\_person\_100007846>} \\
16 & \texttt{<graduatedFrom>}      & rdfs:domain \texttt{<wordnet\_person\_100007846>}   & rdfs:range \texttt{<wordnet\_university\_108286569>} \\
17 & \texttt{<hasChild>}           & rdfs:domain \texttt{<wordnet\_person\_100007846>}   & rdfs:range \texttt{<wordnet\_person\_100007846>} \\
18 & \texttt{<edited>}             & rdfs:domain \texttt{<wordnet\_editor\_110044879>}   & rdfs:range owl:Thing \\
19 & \texttt{<directed>}           & rdfs:domain \texttt{<wordnet\_person\_100007846>}   & rdfs:range \texttt{<wordnet\_movie\_106613686>} \\
20 & \texttt{<wroteMusicFor>}      & rdfs:domain \texttt{<wordnet\_person\_100007846>}   & rdfs:range \texttt{<wordnet\_movie\_106613686>} \\
21 & \texttt{<playsFor>}           & rdfs:domain \texttt{<wordnet\_person\_100007846>}   & rdfs:range \texttt{<wordnet\_organization\_108008335>} \\
22 & \texttt{<isPoliticianOf>}     & rdfs:domain \texttt{<wordnet\_person\_100007846>}   & rdfs:range \texttt{<wordnet\_organization\_108008335>} \\
23 & \texttt{<isLeaderOf>}         & rdfs:domain \texttt{<wordnet\_person\_100007846>}   & rdfs:range \texttt{<wordnet\_organization\_108008335>} \\
24 & \texttt{<influences>}         & rdfs:domain \texttt{<wordnet\_person\_100007846>}   & rdfs:range \texttt{<wordnet\_person\_100007846>} \\
25 & \texttt{<isMarriedTo>}        & rdfs:domain \texttt{<wordnet\_person\_100007846>}   & rdfs:range \texttt{<wordnet\_person\_100007846>} \\
26 & \texttt{<worksAt>}            & rdfs:domain \texttt{<wordnet\_person\_100007846>}   & rdfs:range \texttt{<wordnet\_organization\_108008335>} \\
27 & \texttt{<isInterestedIn>}     & rdfs:domain \texttt{<wordnet\_person\_100007846>}   & rdfs:range owl:Thing \\
28 & \texttt{<livesIn>}            & rdfs:domain \texttt{<yagoLegalActorGeo>}            & rdfs:range \texttt{<wordnet\_location\_100021767>} \\
29 & \texttt{<isKnownFor>}         & rdfs:domain \texttt{<wordnet\_person\_100007846>}   & rdfs:range owl:Thing \\
30 & \texttt{<actedIn>}            & rdfs:domain \texttt{<wordnet\_location\_100021767>}  & rdfs:range \texttt{<wordnet\_movie\_106613686>} \\
31 & \texttt{<hasArea>}            & rdfs:domain \texttt{<wordnet\_location\_100021767>}  & rdfs:range xsd:km2 \\
32 & \texttt{<hasCurrency>}        & rdfs:domain \texttt{<wordnet\_location\_100021767>}  & rdfs:range \texttt{<wordnet\_currency\_108524613>} \\
33 & \texttt{<dealsWith>}          & rdfs:domain \texttt{<wordnet\_person\_100007846>}   & rdfs:range \texttt{<wordnet\_country\_108544813>} \\
34 & \texttt{<hasOfficialLanguage>}      & rdfs:domain \texttt{<wordnet\_location\_100021767>}   & rdfs:range \texttt{<wordnet\_language\_106282651>} \\
35 & \texttt{<hasCapital>}         & rdfs:domain \texttt{<wordnet\_location\_100027167>}     & rdfs:range \texttt{<wordnet\_city\_108524735>} \\
36 & \texttt{<wasCreatedOnDate>}    & rdfs:domain owl:Thing     & rdfs:range xsd:date \\
37 & \texttt{<isLocatedIn>}         & rdfs:domain \texttt{<yagoPermanentlyLocatedEntity>}   & rdfs:range \texttt{<yagoGeoEntity>} \\
38 & \texttt{<hasLongitude>}        & rdfs:domain \texttt{<yagoGeoEntity>}   & rdfs:range \texttt{<degrees>} \\
39 & \texttt{<happenedOnDate>}      & rdfs:domain \texttt{<wordnet\_event\_100029378>}   & rdfs:range xsd:date \\
40 & \texttt{<happenedIn>}         & rdfs:domain \texttt{<wordnet\_event\_100029378>}  & rdfs:range \texttt{<yagoGeoEntity>} \\
41 & \texttt{<hasLatitude>}         & rdfs:domain \texttt{<yagoGeoEntity>}  & rdfs:range \texttt{<degrees>} \\
42 & \texttt{<wasBornIn>}         & rdfs:domain \texttt{<wordnet\_person\_100007846>}  & rdfs:range \texttt{<yagoGeoEntity>} \\
43 & \texttt{<diedIn>}         & rdfs:domain \texttt{<wordnet\_person\_100007846>}  & rdfs:range \texttt{<yagoGeoEntity>} \\
\hline
\end{tabular}
}
\caption{Summary of domain and range properties for selected relations in $\text{S}^2\text{RKE}$.}
\end{table*}

\begin{figure}[t]
  \centering
  \includegraphics[width=0.8\columnwidth]{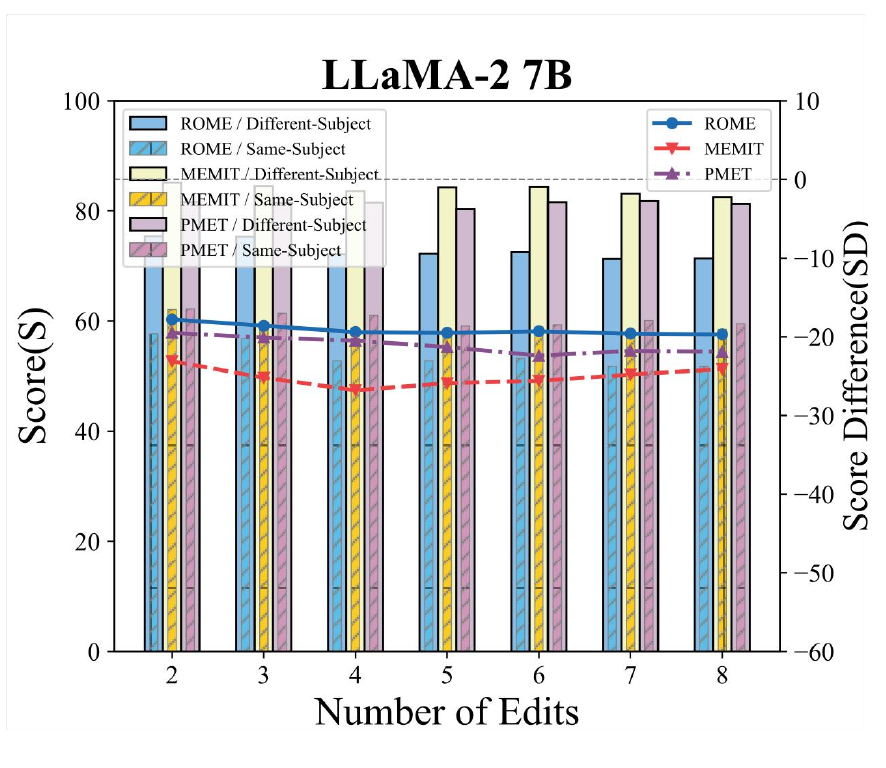}
  \caption{The results of \textit{sequential-editing} on LLaMA-2 7B using mainstream locate-then-edit methods. The bars represent the \textbf{Score (S)} of two strategies, and the line represents the \textbf{Score Difference (SD)} between the two strategies.}
  \label{fig:defference edit 2}
\end{figure}


\begin{figure*}[t]
  \centering
  \begin{subfigure}{2\columnwidth}
    \includegraphics[width=\textwidth]{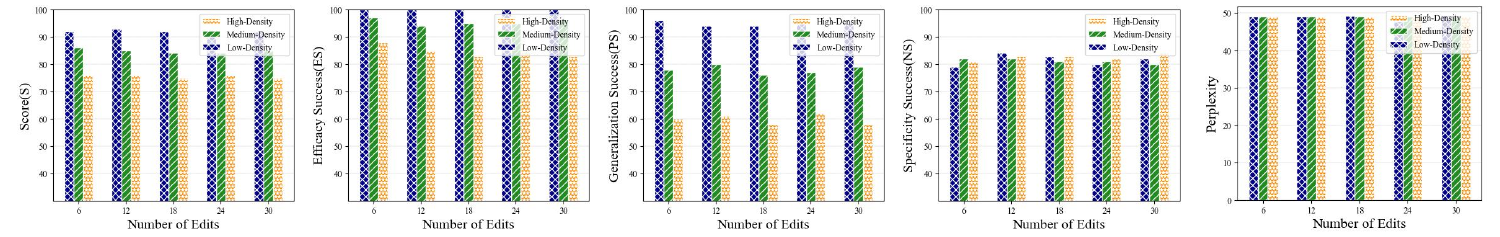}
    \caption{The results of \textit{batch-editing} on GPT-J using MEMIT, comparing five evaluation metrics of three different schemes. }
    \label{fig: pilot batch}
  \end{subfigure}
  
  \vspace{2em} 

  \begin{subfigure}{2\columnwidth}
    \includegraphics[width=\textwidth]{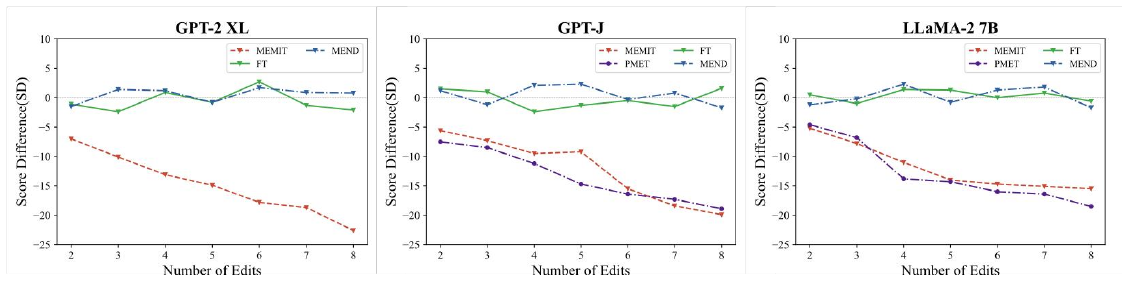}
    \caption{The results of \textit{batch-editing} on three LLMs by six editing methods. \textbf{Score Difference (SD)} represents the difference in editing performance between the two experimental schemes when editing the same amount of knowledge under the same method.}
    \label{fig:Failue batch}
  \end{subfigure}
  
  \vspace{2em} 

  \begin{subfigure}{2\columnwidth}
    \includegraphics[width=\textwidth]{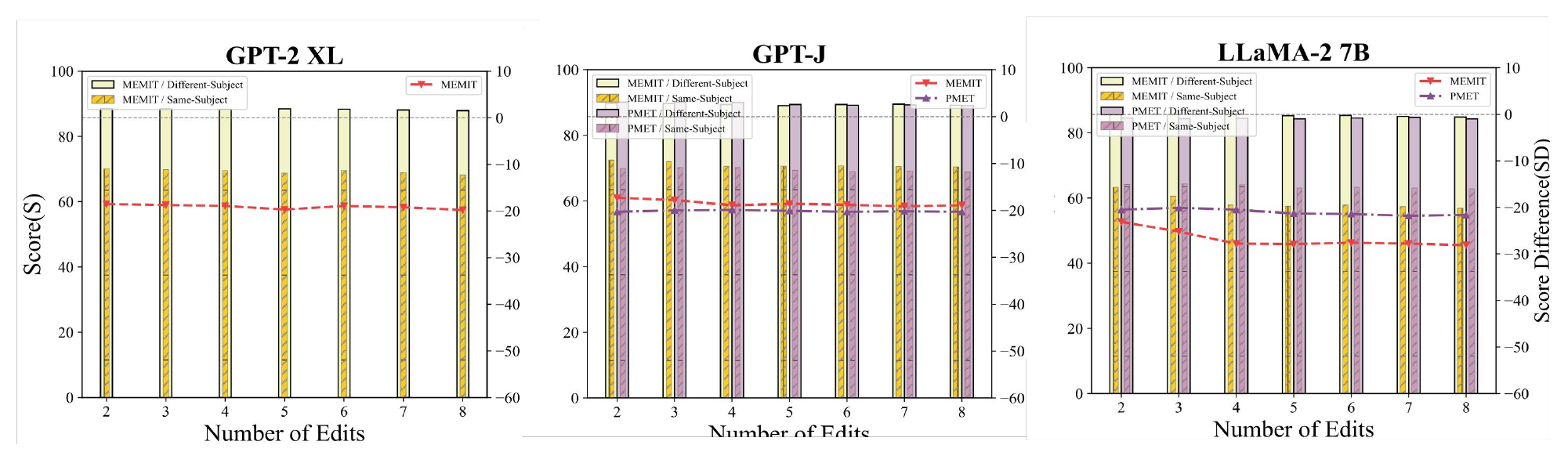}
    \caption{The results of \textit{batch-editing} on three LLMs using mainstream locate-then-edit methods. The bars represent the \textbf{Score (S)} of two strategies, and the line represents the \textbf{Score Difference (SD)} between the two strategies.}
    \label{fig: Failue of defference batch}
  \end{subfigure}
\end{figure*}

\subsection{Data Summary}

\textbf{Data standardization.}\quad Firstly, we standardize the description of each edited to ensure clear distinctions between them. Additionally, we handle relations involving literal- and date-type appropriately, with literal-type storing integers and date-type limited to years. Special characters in object values are also replaced or removed to ensure consistency and operability of the data format.

\textbf{Data statistics.}\quad The $\text{S}^2\text{RKE}$ benchmark contains 6 categories of edited entity, with a total of 3704 subjects and 43 specific relationships, spread across 3 categories of relationship. On average, each entity contains 4.9 edited knowledge entries, with Person entities having the highest number of edits. See Table \ref{table:MAPKE} for statistics of $\text{S}^2\text{RKE}$.

\textbf{Data format.}\quad In summary, each record in the $\text{S}^2\text{RKE}$ benchmark $D$ consists of a subject $s$ and its multiple related requested rewrite. ${r,o, o_{\ast}, P(r)}$. For each rewrite, the benchmark also includes one paraphrase prompt$P^P$ and two neighborhood prompts $P^N$. See Figure for a sample record in SMRKE, complete with three related edits for the same subject.

\begin{figure*}[t]
  \centering
  \includegraphics[width=2.2\columnwidth]{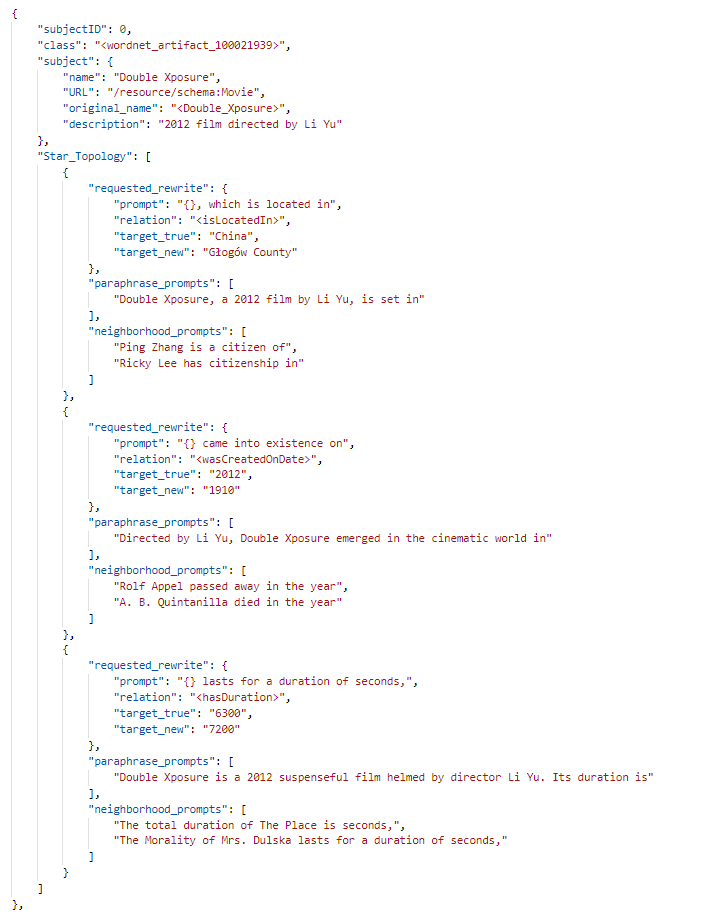}
  \caption{Case example in $\text{S}^2\text{RKE}$.}
  \label{fig: Failue of ROME-like batch}
\end{figure*}

\section{Detailed Experimental Setup} \label{detail experiment}

\subsection{Editing Methods}

In this paper, we use six editing methods: 

\textbf{FT} \cite{zhu2021modifying} applies an $\ell_{\infty}$ norm constraint on the fine-tuning loss, limiting the difference between the original and edited model’s parameters to reduce side effects.

\textbf{MEND} \cite{mitchell2022fast} uses a collection of small hypernetworks to learn a rank-one decomposition of the gradient obtained by standard fine-tuning, enabling tractable edits in LLMs.

\textbf{KN} \cite{dai-etal-2022-knowledge} select neurons associated with knowledge expression via gradient-based attributions, then modify MLP layer at the rows corresponding to those neurons by adding scaled embedding vectors.

\textbf{ROME} \cite{meng2022locating} uses causal tracing to localize the knowledge storage at a specific MLP layer in a transformer, and then updates knowledge by altering the weight matrix with rank-one update.

\textbf{MEMIT} \cite{meng2022memit} extends ROME by distributing updates across multiple MLP layers, enabling large-scale edits.

\textbf{PMET} \cite{li2024pmet} enhances MEMIT by integrating information from both the multi-head self-attention (MHSA) and MLP modules during the optimization process.

It is worth noting that ROME and KN can only \textit{sequential-editing}. All experiments are conducted using the EasyEdit \cite{wang2023easyedit}, ensuring standardized and reproducible evaluations.

\subsection{Selected Models}

In this paper, we select three large language models (LLMs): 

\textbf{GPT-2 XL} \cite{radford2019language}, a 1.5 billion parameter version of GPT-2,is a transformer-based language model developed by OpenAI.

\textbf{GPT-J} \cite{wang2021gpt}, developed by EleutherAI, is a GPT-3-like open-source LLM with 6 billion parameters, trained on \textit{The Pile}.

\textbf{LLaMA2-7B} \cite{touvron2023llama}, a 7 billion parameter version of LLaMA 2 from Meta AI, is a leading open-source LLM, known for its advanced training techniques and optimizations.

\subsection{Evaluation Metrics} \label{Evolution metrics}

To comprehensively evaluate the experimental results, we evaluate editing methods across four dimensions:

\textbf{Efficacy.} We measure efficacy using the Efficacy Success (\textbf{ES}) metric. Specifically, when triple $(s, r, o)$ is updated to $(s, r, o_{\ast})$, ES calculates the success rate of the target edit by determining the probability that the condition $P[o{\ast}] > P[o]$ is satisfied.

\textbf{Generalization.} To evaluate generalization, we use Paraphrase Success (\textbf{PS}) metric, which measures the probability that $P[o_{\ast}] > P[o]$ when the model is prompted with a paraphrase of the original $(s, r)$. 

\textbf{Specificity.}  For specificity, we adopt the Neighborhood Success (\textbf{NS}) metric, which tests the probability that $P[o_c] > P[o_{\ast}]$ for triplet $(s, r, o_c)$, where $o_c$ lies outside the range of the factual edits.

\textbf{Overall Performance.} We assess overall model performance using Perplexity (\textbf{PPL}), based on prior studies by \citet{yang2024butterfly, yang2024fall}. An increase in perplexity generally indicates a decrease in the model's performance in generation tasks.

Finally, to evaluate the balance between efficacy, generalization, and specificity, we report the harmonic mean of \textbf{ES}, \textbf{PS}, and \textbf{NS} indicators as a comprehensive score (\textbf{S}), providing a holistic view of the model’s behavior across these dimensions.

\end{document}